\documentclass[letterpaper, 10 pt,journal, twoside]{IEEEtran}  
\pdfoutput=1

\IEEEoverridecommandlockouts                              

\usepackage{graphics} 
\usepackage{epsfig} 
\usepackage{times} 
\usepackage{amsmath} 
\usepackage{amssymb}  
\usepackage{amsfonts}
\usepackage[colorlinks,linkcolor=blue,anchorcolor=blue,citecolor=blue]{hyperref}
\usepackage[table]{xcolor}
\usepackage{hyperref}
\usepackage{booktabs}
\usepackage{multirow}
\usepackage{graphicx}
\usepackage{subfigure}
\usepackage{cite}
\usepackage{float}
\usepackage{makecell}
\usepackage{algorithm}
\usepackage{colortbl}
\usepackage{hyperref}
\usepackage[noend]{algorithmic}
\usepackage{etoolbox}
\usepackage[flushleft]{threeparttable}
\hypersetup{
    colorlinks=true,
    citecolor=blue    
}

\makeatletter
\patchcmd{\@makecaption}
  {\scshape}
  {}
  {}
  {}
\makeatletter
\patchcmd{\@makecaption}
  {\\}
  {.\ }
  {}
  {}
\makeatother
\def\tablename{Table}

\graphicspath{{img/}}

\usepackage{makecell} 

\title{ARTEMIS: Autoregressive End-to-End Trajectory Planning with Mixture of Experts for Autonomous Driving}

\author{Renju Feng$^\dagger$, Ning Xi$^\dagger$, Duanfeng Chu$^\ast$, Rukang Wang, Zejian Deng,\\ Anzheng Wang, Liping Lu, Jinxiang Wang, Yanjun Huang
\thanks{This work is supported in part by the National Natural Science Foundation of China (52472438), the Key R\&D Program of Hubei Province (2024BAB033), Wuhan Science and Technology Major Project (2022013702025184).}

\thanks{Renju Feng, Ning Xi, Duanfeng Chu,  Rukang Wang , and Anzheng Wang are with the Intelligent Transportation Systems Research Center, Wuhan University of Technology, Wuhan 430063, China (e-mail:{
\href{mailto:fengrenju@whut.edu.cn}{\textcolor{black}{\tt\small fengrenju}}; 
\href{mailto:xining095@whut.edu.cn}{\textcolor{black}{\tt\small xining095}}; 
\href{mailto:chudf@whut.edu.cn}{\textcolor{black}{\tt\small chudf}}; 
 \href{mailto:wangrk@whut.edu.cn}{\textcolor{black}{\tt\small wangrk}}; \href{mailto:wanganzheng@whut.edu.cn}{\textcolor{black}{\tt\small wanganzheng@whut.edu.cn}}}).}
\thanks{Zejian Deng is with the Department of Data and Systems Engineering, The University of Hong Kong, Hong Kong SAR, 999077, China (e-mail: {\tt\small \href{mailto:z49deng@hku.hk}{\textcolor{black}{\tt\small z49deng@hku.hk}}}).}
\thanks{Liping Lu is with the School of Computer Science and Artificial Intelligence, Wuhan University of Technology, Wuhan 430070, China (e-mail: {\tt\small \href{mailto:luliping@whut.edu.cn}{\textcolor{black}{\tt\small luliping@whut.edu.cn}}}).}
\thanks{Jinxiang Wang is with the School of Mechanical Engineering, Southeast University, Nanjing 211189, China (e-mail: {\tt\small wangjx@seu.edu.cn}).}
\thanks{Yanjun Huang is with the School of Automotive Studies, Tongji University,
Shanghai 201804, China (e-mail: {\tt\small yanjun\_huang@tongji.edu.cn}).}
\thanks{$^\dagger$Equal contribution.}
\thanks{$^\ast$Corresponding author.}
}

\providecommand{\keywords}[1]{\textbf{\textit{Index terms---}} #1}
\begin{document}

\maketitle
\pagestyle{empty} 
\thispagestyle{empty} 

\begin{abstract}This paper presents ARTEMIS, an end-to-end autonomous driving framework that combines autoregressive trajectory planning with Mixture-of-Experts (MoE). 
Traditional modular methods suffer from error propagation, while existing end-to-end models typically employ static one-shot inference paradigms that inadequately capture the dynamic changes of the environment. 
ARTEMIS takes a different method by generating trajectory waypoints sequentially, preserves critical temporal dependencies while dynamically routing scene-specific queries to specialized expert networks. 
It effectively relieves trajectory quality degradation issues encountered when guidance information is ambiguous, and overcomes the inherent representational limitations of singular network architectures when processing diverse driving scenarios.
Additionally, we use a lightweight batch reallocation strategy that significantly improves the training speed of the Mixture-of-Experts model. Through experiments on the NAVSIM dataset,
ARTEMIS exhibits superior competitive performance, achieving 87.0 PDMS and 83.1 EPDMS with ResNet-34 backbone, demonstrates state-of-the-art performance on multiple metrics.
Code will be available under \url{https://github.com/Lg0914/ARTEMIS}.
\end{abstract}

\section{Introduction}
\begin{figure}
    \centering
    \includegraphics[width=\columnwidth]{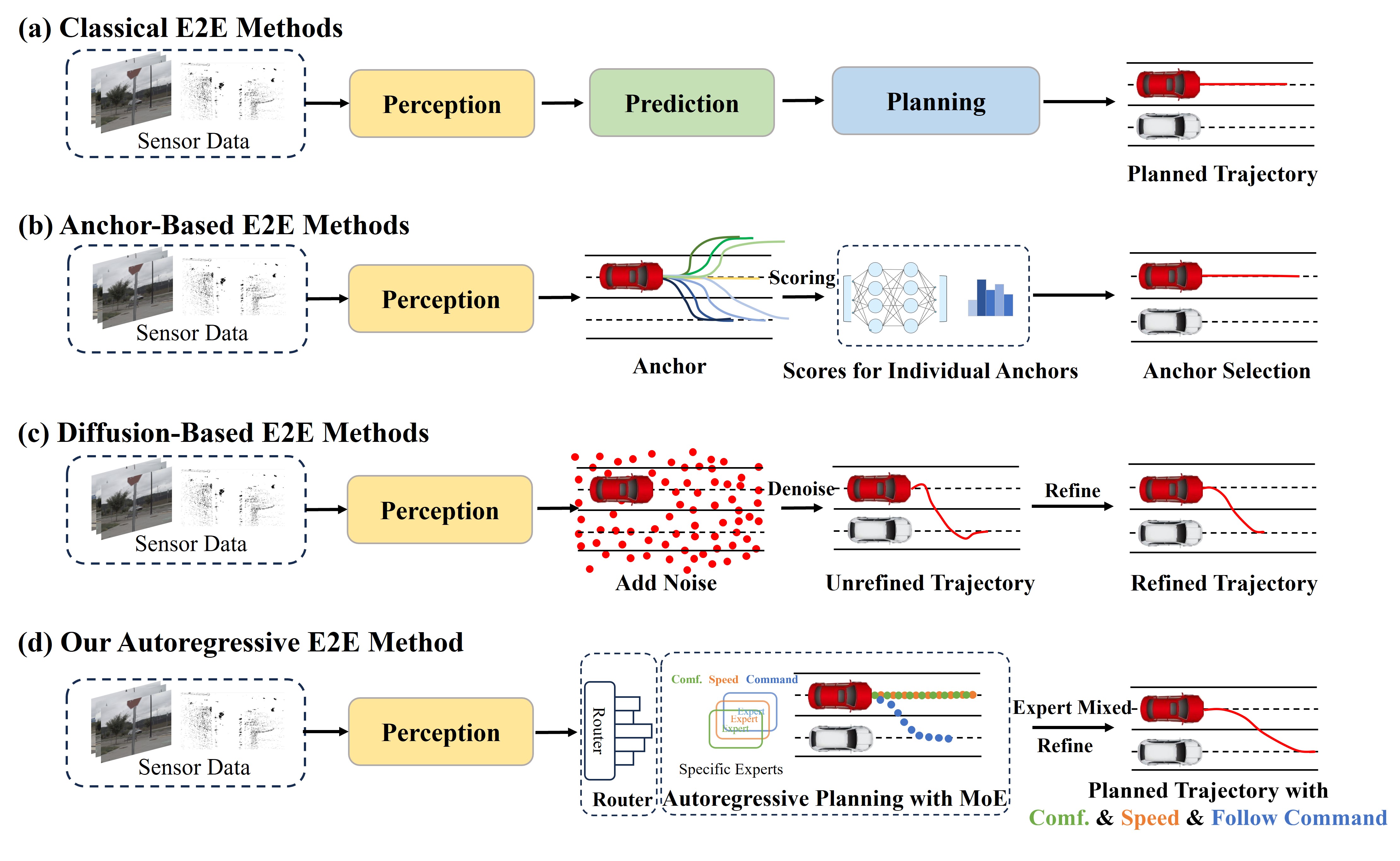}
    \vspace{-0.5cm}
    \caption{\textbf{Four different end-to-end architectures.} (a) Classical end-to-end methods\cite{hu2023uniad,jiang2023vad,yuan2024drama,chitta2022transfuser}. (b) Anchor-based end-to-end methods\cite{chen2024vadv2,li2024hydra,li2025hydra++}. (c) Diffusion-based end-to-end methods\cite{liao2024diffusiondrive}. (d) Our proposed method.}
    \vspace{-0.5cm}
    \label{fig:introduction}
\end{figure}

\IEEEPARstart{A}{utonomous} driving has experienced rapid development over the past few decades. 
Traditional modular methods compartmentalize autonomous driving tasks into discrete modules such as perception, prediction, and planning\cite{li2024bevformer,xin2025multi,cheng2024pluto}.
However, cumulative errors and complex interdependencies between these modules can become constrained at predefined interfaces. 
End-to-end models overcome these issues by mapping raw sensor data directly to planned trajectories or control signals\cite{hu2023uniad,jiang2023vad,wu2022tcp}. 
However, their static, one-shot inference paradigm often fails to capture the dynamic evolution of the environment (Fig.\ref{fig:introduction} (a)). 
In contrast, autoregressive methods generate trajectories sequentially, preserving temporal coherence and allowing for adaptive decision-making based on previously planned segments.
Autoregressive models have been widely applied in the field of trajectory prediction\cite{liu2021multimodal,yuan2021agentformer,rhinehart2018r2p2}. 
Recent research has begun to explore unified frameworks that simultaneously accomplish world model construction and complete or partial trajectory planning tasks through autoregressive modeling methods\cite{chen2024drivinggpt,hu2024drivingworld}. 
However, the current problem is that single-network end-to-end models still struggle to adequately capture and adapt to diverse driving scenarios.

To tackle the inherent complexity in trajectory planning, researchers are increasingly adopting sophisticated architectures like the Mixture of Experts (MoE) framework\cite{jacobs1991adaptive}. 
MoE employs multiple specialized expert networks along with intelligent routing mechanisms to dynamically distribute and process inputs—a strategy that has yielded significant success in large-scale language models\cite{dai2024deepseekmoe,achiam2023gpt,fedus2022switch}. 
In autonomous driving, the planning trajectories generated by end-to-end models inherently encompass multiple potential behavioral modalities, reflecting the fundamental uncertainty in driving behavior. 
Drivers may select from several reasonable future actions under identical environmental conditions (Fig.\ref{fig:introduction} (b))\cite{chen2024vadv2,li2024hydra,li2025hydra++}, and traditional single-network architectures struggle to accurately characterize this intrinsic diversity of behavioral patterns. 
In contrast, MoE enables expert modules to focus on specific driving scenarios or behavioral patterns, thereby learning characteristic distributions of driving behavior without reliance on predefined guidance signals. 
This endogenous multimodal modeling method effectively circumvents potential trajectory quality degradation issues that may arise when guidance information deviates from actual conditions. 
Previous work, which was based on structured data representations as input\cite{sun2024str}, incorporated Mixture-of-Experts (MoE) into planning tasks and demonstrated strong performance on the NuPlan dataset\cite{caesar2021nuplan}.
Recently, diffusion-based models (Fig.\ref{fig:introduction} (c))\cite{liao2024diffusiondrive}, \cite{janner2022planning} have introduced novel generative modeling paradigms to autonomous driving, demonstrating trajectory diversity. 
Although these methodologies have established state-of-the-art performance in end-to-end autonomous driving, they typically employ static paradigms that generate all trajectory points simultaneously (or through multiple denoising iterations), limiting their capacity to accurately capture the dynamic evolutionary characteristics of trajectory development.
In comparison, autoregressive methods utilizing MoE demonstrate superior temporal sequence capturing capabilities, environmental adaptability, and practical utility, as they can operate without requiring strong prior constraints.

\begin{figure*}
\centering

\includegraphics[width=1\textwidth]{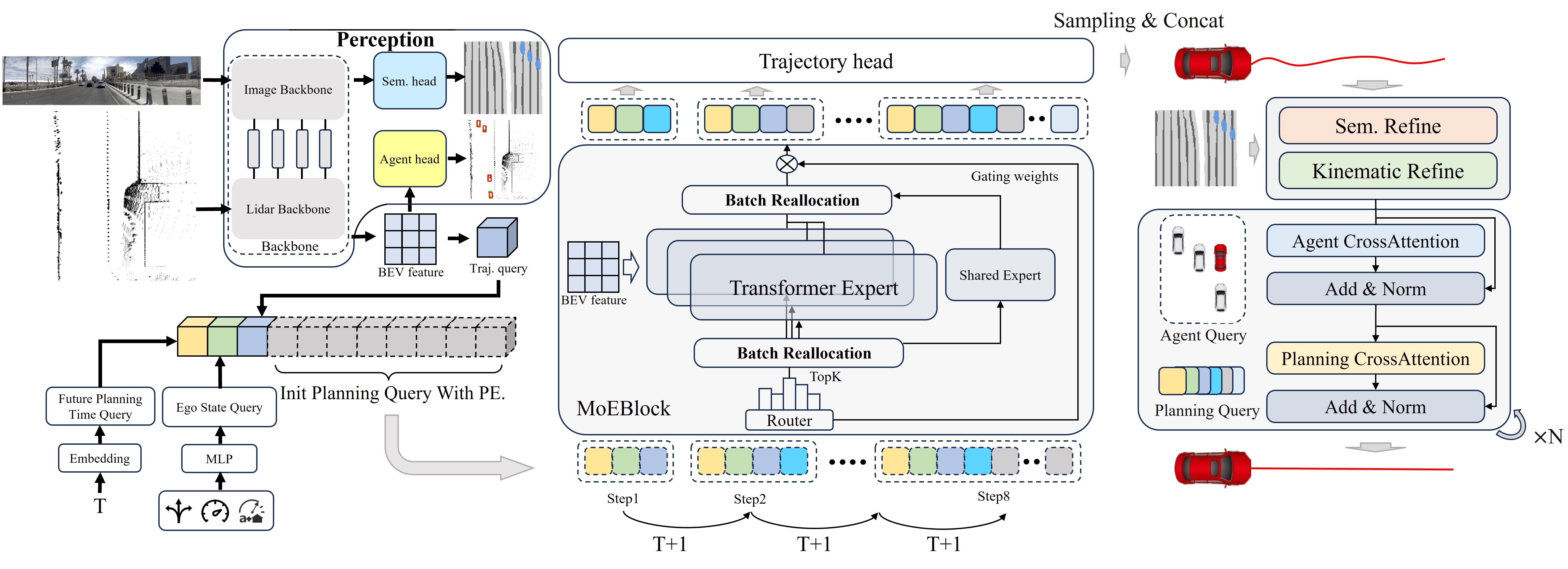}
\vspace{-0.5cm}
\caption{The figure illustrates our method: The perception module utilizes separate backbone networks for image and LiDAR data, ultimately fusing them into BEV feature representations. The autoregressive planning module with MOE progressively generates trajectory waypoints through sequential decision processes while dynamically selecting the most appropriate expert network for the current environment. Additionally, it implements batch reallocation based on expert activation patterns. Finally, the trajectory refinement module processes and refines the autoregressive trajectory outputs.}
\vspace{-0.3cm}
\label{fig:framework}
\end{figure*}

To address these issues, we propose ARTEMIS, Autoregressive End-to-End Trajectory Planning with Mixture of Experts for Autonomous Driving, as shown in Fig.\ref{fig:framework}. 
ARTEMIS comprises three primary components: a perception module, an autoregressive planning module with MOE, and a trajectory refinement module. 
The perception module adopts the Transfuser\cite{chitta2022transfuser}, utilizing separate backbone networks for image and LiDAR data, ultimately fusing them into BEV feature representations. 
The autoregressive planning module with MOE progressively generates trajectory waypoints through sequential decision processes while dynamically selecting the most appropriate expert network for the current driving scenario. 
Additionally, it implements batch reallocation based on expert activation patterns. 
Finally, the trajectory refinement module processes and refines the autoregressive trajectory outputs.
We conducted comprehensive evaluations of ARTEMIS using the NAVSIM dataset\cite{dauner2024navsim}. 
Our contributions can be summarized as follows:

(1) To the best of our knowledge, this study represents the first investigation to incorporate the Mixture-of-Experts (MoE) into end-to-end autonomous driving, effectively relieving trajectory quality degradation issues encountered in traditional methods when guidance information is ambiguous, as well as the inherent representational limitations of singular network architectures when processing diverse driving scenarios, through dynamic routing mechanisms and specialized expert network division.

(2) We propose an autoregressive end-to-end planning methodology that constructs trajectories sequentially through iterative decision processes, enabling precise modeling of strong temporal dependencies between trajectory waypoints.

(3) Our method achieves significant results on the large-scale real-world NAVSIM dataset. Using an identical ResNet-34 backbone, our method attains 87.0 PDMS under standard metrics and 83.1 EPDMS under extended evaluation metrics.

\section{Method}
\label{sec:result}
\subsection{Preliminaries}
\subsubsection{End-to-End Autonomous Driving}

Formally, we define end-to-end autonomous driving as a conditional sequence generation problem where the input is a sequence of historical sensor observations $S=(s_1,s_2,...,s_T)$, where $T$ denotes the history horizon. 
The model is required to generate a trajectory sequence for future time instants $Y=(y_1,y_2,...,y_H)$, where $H$ denotes the prediction horizon. 
Each point $y_h \in\mathcal{R}^{d}$, with $d$ representing the dimensionality of each waypoint. 
The objective of an end-to-end model is to learn the conditional probability distribution $p(Y|S)$.

For autoregressive models, we can decompose the conditional probability distribution as follows:
\begin{equation}
p(Y|S)=\prod_{t=1}^Hp\left(y_t|y_{<t},S\right)
\end{equation}
where \(y_{<t}\) denotes trajectory points generated prior to time \(t\). 

In a typical end-to-end model, the sensor data \(S\) is first transformed into a latent representation \(z = \phi(S)\) via a feature extraction network \(\phi(S)\), and then mapped to the final trajectory \(Y\) using a trajectory generation network $\psi()$.

\subsubsection{Mixture of Experts}
In this work, we adopt a design similar to DeepseekMoE~\cite{dai2024deepseekmoe}, where some experts are designated as shared experts to capture generalizable knowledge and reduce the redundancy of routing experts. 
Given an input $x$, the output of our MoE can be formalized as:
\begin{equation}
y=\sum_{i\in E_{shared}}f_i(x)+\sum_{i\in E_{private}}g_i\left(x\right)\cdot f_i(x)
\end{equation}
where \(E_{\text{shared}}\) represents the shared experts, \(E_{\text{private}}\) denotes the domain-specific experts, and \(E\) denotes the total number of expert networks. 
The computational function of the \(i\)th expert is given by $f_i: \mathbb{R}^{n \times d_{\text{model}}}$, 
and the gated neural network assigns a weight to the \(i\)th expert via the function $g_i: \mathbb{R}^{n \times d_{\text{model}}} \to [0,1]$.

\subsection{Model Architecture}
\subsubsection{Perception Module}

The perception module adheres to the design of the Transfuser~\cite{chitta2022transfuser} and is responsible for extracting features from raw sensor data. 
This module employs a multimodal fusion strategy to process both image and point cloud data simultaneously, thereby constructing a unified representation of the environment. 
Specifically, the module consists of two parallel feature extractors with feature fusion conducted at various stages through a Transformer. 

The module utilizes point cloud data $D_l \in \mathbb{R}^{256 \times 256}$ and front-view image data $D_I \in \mathbb{R}^{1024 \times 256 \times 3}$, from which visual features (e.g., \(F_{cam}\) and \(F_{lidar}\)) are extracted using a series of convolution layers and a ResNet-34 backbone. Finally, a multimodal fusion mechanism is employed to integrate these features into a Bird's-Eye View (BEV) feature representation $F_{\text{bev}} \in \mathbb{R}^{B \times C_{\text{bev}} \times d_{\text{model}}}$.

\subsubsection{Autoregressive Planning Module with Mixture-of-Experts}

Unlike traditional one-shot methods, this study employs an autoregressive strategy to incrementally build trajectories while integrating a Mixture-of-Experts (MoE) architecture. 
This design leverages both previous trajectory information and dynamically selected specialized expert networks based on scene characteristics. 

\begin{figure}
    \centering
    \includegraphics[width=\columnwidth]{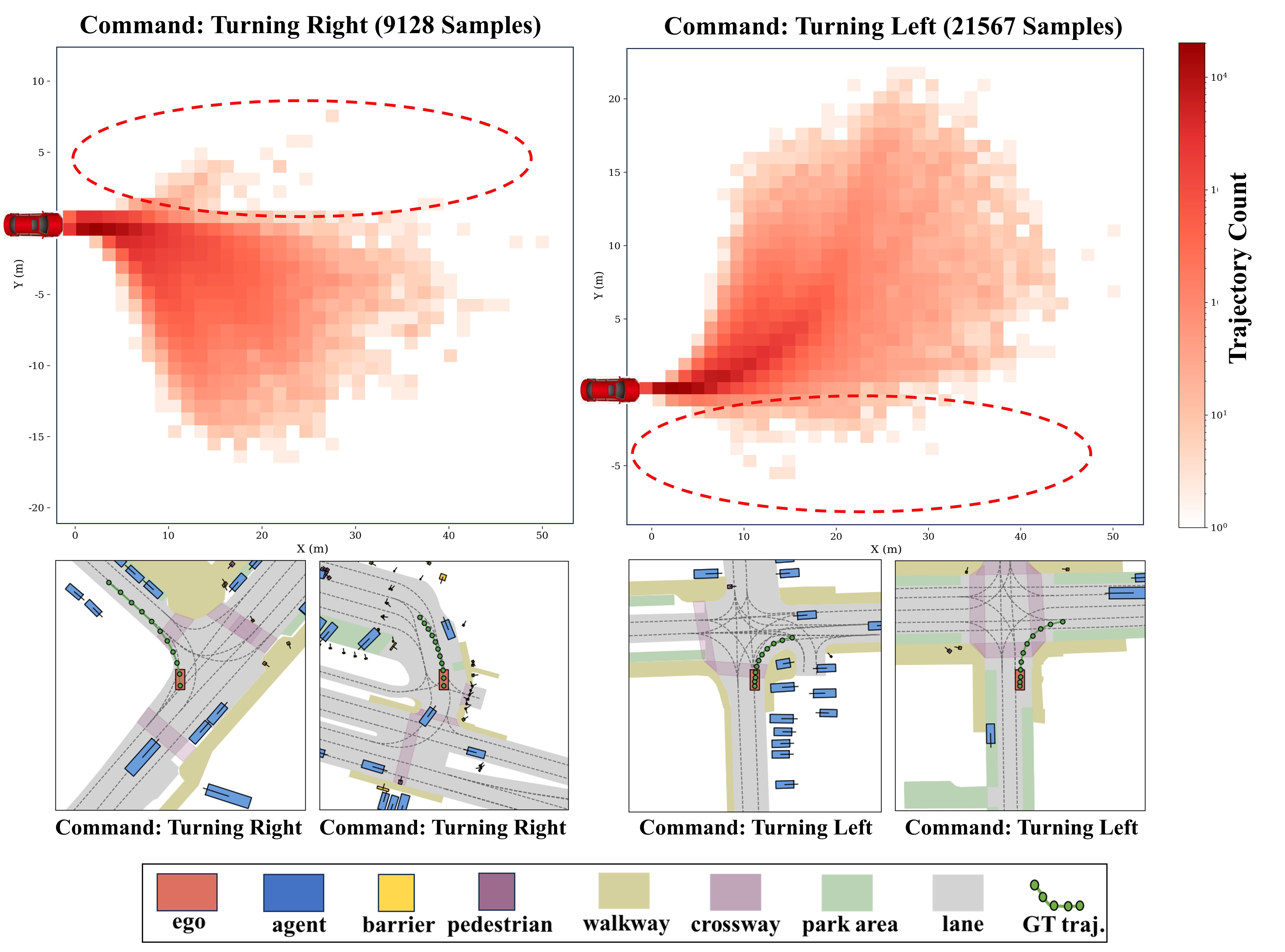}
    \vspace{-0.7cm}
    \caption{The lower part of this figure shows several typical scenes from the navtrain split where control commands are inconsistent with the actual expert driving trajectories. The upper part visualizes the distribution of expert trajectories under right and left turn control commands through grid heatmaps, clearly showing trajectory deviations where some grids are inappropriately activated due to incorrect control commands.}
    \label{fig:navtrain}
    \vspace{-0.5cm}
\end{figure}

Our analysis of the navtrain split reveals a significant imbalance in the distribution of driving commands (with over 20,000 samples for left-turn commands, fewer than 10,000 samples for right-turn commands, and more than 50,000 samples for forward driving commands). 
\textbf{Furthermore, we observed that a small proportion of the training samples exhibit discrepancies between the driving commands and the expert trajectories} (as illustrated in Fig.\ref{fig:navtrain}). 
Relying solely on driving commands for expert selection may result in insufficient training data for certain experts and fail to capture the diversity of driving strategies. 
Ultimately, we adopt an endogenous routing multimodal modeling method to effectively relieve these challenges.

To avoid the model exhibiting a bias toward directly learning planning from historical ego trajectory, which may lead to causal confusion and other issues~\cite{li2024ego, chu2024}. 
We exclusively encode the current ego state $s_0 \in \mathbb{R}^{8}$ (including control commands, 2D velocity, and acceleration) into a feature space $Q_s$ via an MLP.

\textbf{Positional and Temporal Embedding.} It is essential to inform the model of the specific time step for which the trajectory point is being planned. 
For a future planning time step \(t\), we employ an embedding layer to obtain the planning time embedding, denoted as \(TE_t\). 
Similarly, we utilize a positional embedding to incorporate positional information into the planning sequence, represented by \(PE_t\). 
It is important to note that we only add the positional embedding to the initialized planning sequence during the first autoregressive step, thus avoiding the accumulation of extraneous noise that may result from repeatedly adding the embedding information.

\textbf{Autoregressive Generation.} We first add positional embeddings to the complete current planning sequence for initialization. 
This sequence is then fed into a Transformer encoder with a padding mask \(M_t\) to update the planning queries $Q_{1:t}$. The padding mask ensures that the planning query at the current time step interacts only with the historical planning queries in the sequence.
By constructing a concatenated query that integrates the planning time information, ego state, and the current as well as historical planning queries, we have $C_t = \text{Concat}(TE_t, Q_s, Q_{(1:t)})$. 
This concatenated query is then fed into the MoEBlock with batch reallocation to interact with the BEV feature and obtain the planning query for the current time step.
\begin{equation}
Q_{t+1}=MoEBlock(F_{bev},C_t)
\end{equation}

Finally, to further characterize the inherent uncertainty in driving behavior, we adopt a probabilistic modeling method. 
A multi-layer MLP network is employed to predict the distribution of multimodal trajectory points, encompassing both position and heading. 
Ultimately, a trajectory point at the current planning time step is generated by sampling from the predicted distribution $y_t \sim \mathcal{N}(\mu_t, \sigma_t^2)$.
After the trajectory point is generated, the latest planning query corresponding to the current time step is updated in the planning query sequence.
Finally, the individual trajectory points are concatenated to form the initial trajectory.

\begin{figure}
    \centering
    \includegraphics[width=\columnwidth]{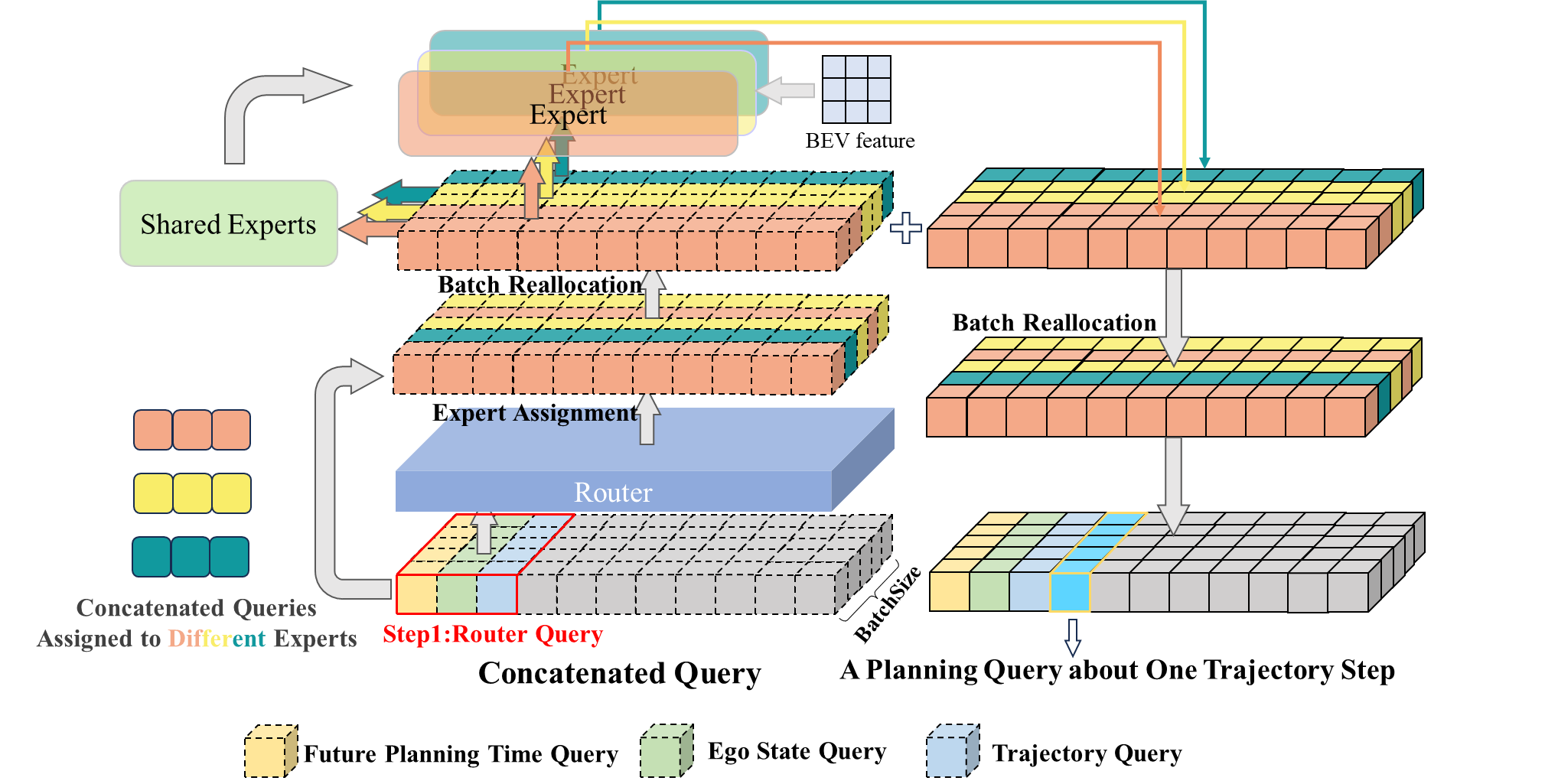}
    \vspace{-0.5cm}
    \caption{\textbf{Batch Reallocation MoE.} This module intelligently allocates expert resources to concatenated queries from different samples through the routing network, and employs batch reorganization to group sample data that activate the same experts together, thereby improving computational efficiency. The system subsequently uses shared expert networks and different specialized expert networks to process and fuse BEV feature, then restores the original sample order through inverse batch reorganization after processing. Finally, the module extracts the query corresponding to the next position from the processed concatenated query, which serves as the planning query input for the next trajectory point supplied to the trajectory planning head.}
    \vspace{-0.5cm}

    \label{fig:moe}
\end{figure}

\textbf{Batch Reallocation MoE.} This module, as shown in Fig.\ref{fig:moe}, includes both shared experts \(E_{\text{shared}}\) and domain-specific experts \(E_{\text{private}}\). 
We use an efficient batch reallocation strategy that significantly enhances computational efficiency, particularly when processing large-scale data. 
To ensure a fixed query dimensionality for the routing network, we remove the historical planning queries from the concatenated query, forming the routing query $Q_r = \text{Concat}(TE_t, Q_s, Q_t)$. The router network $\mathcal{R}$ consists of two MLPs and is used to compute the expert assignment scores \(g_t\). 
The first MLP performs dimensionality reduction on input features, while the second maps compressed features to expert scores. 
This two-stage design efficiently balances computational cost with routing decision quality.

To enable the model to focus on the most relevant experts during training, we also adopt a sparse activation strategy by selecting only the top \(k\) experts with the highest score. 
For each selected expert index sequence vector \(a_i\), we perform a series of operations including batch sorting, data reorganization, and block identification. 
Specifically, we first sort the batch samples according to the expert indices, resulting in a sorting function \(\pi_i\). Subsequently, the BEV feature and the concatenated queries are restructured based on this sorting function as follows:
\begin{equation}
F_{\text{bev}}^{(i)} = \pi_i(F_{\text{bev}}) \quad \text{and} \quad Q_t^{(i)} = \pi_i(Q_t).
\end{equation}

Based on expert indexing patterns, consecutive blocks with identical expert indices are identified and matched:
\begin{equation}
\{E_j^i,n_j^i\}_{j=1}^J=UC(\pi_i)
\end{equation}
where $J$ represents the total number of blocks, $E_j^i$ denotes the consecutive block corresponding to the $j_{th}$ expert in the $i_{th}$ expert index sequence vector, and $n_j^i$ represents the corresponding block size as determined by the unique consecutive function. 
The appropriate expert network is applied to each block, where $F_{\text{bev}}^{(i,j)}$ and $C_t^{(i,j)}$ constitute the reorganized expert input data for block $E_j$ under the expert index sequence vector $a_i$.
\begin{equation}
F_{bev}^{(i,j)}=F_{bev}^{(i)}[b_j^i{:}(b_j^i+n_j^i)]
\end{equation}
\begin{equation}
C_t^{(i,j)}=C_t^{(i)}[b_j^i{:}(b_j^i+n_j^i)]
\end{equation}
\begin{equation}
\boldsymbol{O}_{i,j}=\boldsymbol{E}_{private}^{j}(F_{bev}^{(i,j)},\boldsymbol{C}_{t}^{(i,j)})+\boldsymbol{E}_{shared}(\boldsymbol{F}_{bev}^{(i,j)},\boldsymbol{C}_{t}^{(i,j)})+\boldsymbol{C}_{t}^{(i,j)} 
\end{equation}
where $b_j^i$ denotes the starting position of the consecutive block corresponding to the $j_{th}$ expert in the $i_{th}$ expert index sequence vector. 

Finally, the expert processing results are restored to their original batch ordering, and outputs from different experts are fused according to their corresponding weights $g_{ij}$ to derive the final composite output $O_f$.
\begin{equation}
\pi_i(\widehat{\boldsymbol{O}}_{i,j})=\boldsymbol{O}_{i,j}\quad \text{and} \quad O_f=\sum_{i}\sum_{j}(g_{ij}\times \widehat{\boldsymbol{O}}_{i,j})
\end{equation}

\subsubsection{Trajectory Refinement Module}

Considering that driving scenarios are inherently complex and often contain information that is unrelated to planning or even constitutes noise, we introduce a trajectory refinement module. 
The refinement process ensures that the final trajectory $Y$ satisfies kinematic constraints, avoids obstacles, and maintains smoothness. This process is divided into two stages: semantic kinematic optimization and cross-attention refinement.

\textbf{Semantic Kinematic Optimization.} In the semantic optimization stage, we extract features from the BEV semantic map \(S_{\text{bev}}\) to obtain semantic features $F_{\text{sem}} = \phi_{\text{sem}}(S_{\text{bev}})$,
where \(\phi_{\text{sem}}\) is a semantic encoder that is primarily implemented using a multi-layer convolutional network. 
We encode the initial trajectory using a GRU network to obtain the trajectory features \(h_{\text{traj}}\), and subsequently feed both the semantic features \(F_{\text{sem}}\) and \(h_{\text{traj}}\) into an optimizer network \(\phi_{\text{optimizer}}\),  yielding the joint representation $F_{\text{combined}}$.

The decoding part optimizes the trajectory point by point by fusing the features. 
For each trajectory point, a GRU and an output layer \(\phi_{\text{output}}\) are used to generate the optimized point \(y_i'\). 
In the kinematic optimization stage, multiple explicit constraints are employed to optimize the generated trajectory points, including both smoothness constraints and kinematic constraints, with learnable weights assigned to these constraints. 
The final optimized points obtained from semantic kinematic optimization are denoted as \(\hat{y}_i\).

\textbf{Cross-Attention Refinement.} We utilize a cascaded cross-attention module to further enhance the interaction between the trajectory and the scene context, enabling the trajectory features to interact with both the agent features and the ego planning features.
\begin{equation}
Y=\Psi_{refine}(\hat{y}_i,Q_{agent},Q_{ego})
\end{equation}
where \(Q_{\text{agent}}\) represents the agent query features, \(Q_{\text{ego}}\) denotes the ego planning query features, and \(\Psi_{\text{refine}}\) is the refinement function.

\subsubsection{Training Loss}
Consistent with some end-to-end methods, we adopt a staged training method to mitigate training instability\cite{hu2023uniad,jiang2023vad,chen2024vadv2}. Specifically, the perception network and its auxiliary tasks, including semantic mapping and object detection, are first trained. 
Subsequently, the entire network is trained in an end-to-end method. 
This training strategy significantly enhances both the stability and the overall performance of the model.
Furthermore, we opted not to employ the expert balance loss commonly utilized in MoE architecture, as its application in datasets with imbalanced feature distributions could potentially impede the acquisition of specialized strategic knowledge by individual experts.

\textbf{Perception Stage Loss.} In the first stage, we focus on optimizing the auxiliary tasks related to perception. The total loss is defined as:
\begin{equation}
\mathcal{L}_{perception}=\lambda_{sem}\mathcal{L}_{sem}+\lambda_{class}\mathcal{L}_{class}+\lambda_{box}\mathcal{L}_{box}
\end{equation}
where $\mathcal{L}_{sem}$ is the cross-entropy loss for the BEV semantic map, and $\mathcal{L}_{class}$ and $\mathcal{L}_{box}$ are the agent classification and localization losses computed using the Hungarian matching algorithm.

\textbf{End-to-End Training Loss.} In this stage, the entire network is trained in end-to-end method. The overall loss is defined as:
\begin{equation}
\mathcal{L}_{planning}=\lambda_{traj}\mathcal{L}_{traj}+\lambda_{NLL}\mathcal{L}_{NLL}+\mathcal{L}_{perception}
\end{equation}
where $\mathcal{L}_{traj}$ denotes the planning \(L_1\) loss, and $\mathcal{L}_{NLL}$ represents the negative log-likelihood loss.

\begin{table*}[ht]
  \centering

  \caption{\textbf{Performance on the Navtest Benchmark with Original Metrics.}}
  \label{tab:navsim_v1}
  \begin{tabular}{c|lc|ccccc>{\columncolor{gray!20}}c}
    \toprule
    \textbf{Type}&\textbf{Method}&\textbf{Input} & \textbf{NC$\uparrow$} & \textbf{DAC$\uparrow$} & \textbf{EP$\uparrow$} & \textbf{TTC$\uparrow$} & \textbf{C$\uparrow$} & \textbf{PDMS$\uparrow$} \\
    \midrule
    Diffusion-based &DiffusionDrive~\cite{liao2024diffusiondrive} &C \& L& \underline{98.2} & \textbf{96.2} & \textbf{82.2} & \underline{94.7} & \textbf{100} & \textbf{88.1} \\
    \midrule
    \multirow{10}{*}{IL-based} &ADMLP~\cite{zhai2023rethinking} &-& 93.0 & 77.3 & 62.8 & 83.6 & \textbf{100} & 65.6 \\
    &VADv2~\cite{chen2024vadv2} &C \& L& 97.2 & 89.1 & 76.0 & 91.6 & \textbf{100} & 80.9 \\
    &UniAD~\cite{hu2023uniad} &C& 97.8 & 91.9 & 78.8 & 92.9 & \textbf{100} & 83.4 \\
    &LTF~\cite{chitta2022transfuser}&C & 97.4 & 92.8 & 79.0 & 92.4 & \textbf{100} & 83.8 \\
    &Transfuser~\cite{chitta2022transfuser}&C \& L & 97.7 & 92.8 & 79.2 & 92.8 & \textbf{100} & 84.0 \\
    &PARA‑Drive~\cite{weng2024para} &C& 97.9 & 92.4 & 79.3 & 93.0 & 99.8 & 84.0 \\
    &DRAMA~\cite{yuan2024drama}&C \& L & 98.0 & 93.1 & 80.1 & \textbf{94.8} & \textbf{100} & 85.5 \\
    &GoalFlow$^*$~\cite{xing2025goalflow}&C \& L & \textbf{98.3} & 93.8 & 79.8 & 94.3 & \textbf{100} & 85.7 \\
    &Hydra‑MDP$^*$~\cite{li2024hydra} &C \& L& \textbf{98.3} & \underline{96.0} & 78.7 & 94.6 & \textbf{100} & 86.5 \\
    &Hydra‑MDP++$^*$~\cite{li2025hydra++}&C & 97.6 & \underline{96.0} & 80.4 & 93.1 & \textbf{100} & 86.6 \\
    \midrule
    IL-based &\textbf{Ours}&C \& L & \textbf{98.3} & 95.1 & \underline{81.4} & 94.3 & \textbf{100} & \underline{87.0} \\
    \bottomrule
  \end{tabular}
  \vspace{0.2cm}
  
  \raggedright \footnotesize In the ``Input'' column, ``C'' denotes the use of camera images as sensor input, and ``C\&L'' represents the use of both camera images and lidar as sensor inputs. All models require ego-vehicle state as input. * For fair comparison, we used the official scores of versions with the same backbone network and hidden layer dimensions. The best and second-best scores are highlighted in \textbf{bold} and \underline{underlined}, respectively.
\end{table*}

\begin{table*}[ht]
  \centering
  \vspace{-0.3cm}
  \caption{\textbf{Performance on the Navtest Benchmark with Extended Metrics.}} 
  \label{tab:navsim_v1_1}
  \begin{tabular}{l|ccccccccc>{\columncolor{gray!20}}c}
    \toprule
    \textbf{Method} & \textbf{NC$\uparrow$} & \textbf{DAC$\uparrow$} & \textbf{EP$\uparrow$} &
    \textbf{TTC$\uparrow$} & \textbf{C$\uparrow$} & \textbf{TL$\uparrow$} &
    \textbf{DDC$\uparrow$} & \textbf{LK$\uparrow$} & \textbf{EC$\uparrow$} &
    \textbf{EPDMS$\uparrow$} \\
    \midrule
    Transfuser~\cite{chitta2022transfuser}               & 97.7 & 92.8 & 79.2 & 92.8 & \textbf{100} & \underline{99.9} & 98.3 & \underline{67.6} & 95.3 & 77.8 \\
    VADv2~\cite{chen2024vadv2}                    & 97.3 & 91.7 & 77.6 & 92.7 & \textbf{100} & \underline{99.9} & 98.2 & 66.0 & 97.4 & 76.6 \\
    Hydra‑MDP~\cite{li2024hydra}                & 97.5 & \underline{96.3} & \underline{80.1} & 93.0 & \textbf{100} & \underline{99.9} & 98.3 & 65.5 & 97.4 & 79.8 \\
    Hydra‑MDP++~\cite{li2025hydra++} & \underline{97.9} & \textbf{96.5} & 79.2 & \underline{93.4} & \textbf{100} & \textbf{100} & \textbf{98.9} & 67.2 & \underline{97.7} & \underline{80.6} \\
    \midrule
    Ours & \textbf{98.3} & 95.1 & \textbf{81.5} & \textbf{97.4} & \textbf{100} & 99.8 & \underline{98.6} & \textbf{96.5} & \textbf{98.3} & \textbf{83.1} \\
    \bottomrule
  \end{tabular}
  \vspace{0.2cm}
  
  \raggedright \footnotesize Comparison model results are sourced from Hydra-MDP++~\cite{li2025hydra++}.
\vspace{-0.3cm}
\end{table*}

\section{Experiments}

\subsection{Dataset}
We train and test our model on the NAVSIM dataset~\cite{dauner2024navsim}. 
NAVSIM selects challenging scenarios from the OpenScene dataset\cite{contributors2023openscene}, excluding simple scenarios. 
The training set comprises 1192 scenarios, while the test set contains 136 scenarios. 
Each sample in the dataset includes camera images from 8 viewpoints, LiDAR data fused from 5 sensors, map annotations, and 3D object bounding boxes, among other data.
In the NAVSIM dataset, the model is required to use 4 frames of 2 seconds of historical and current data to plan a 4-second trajectory consisting of 8 future frames.

\vspace{-0.3cm}
\subsection{Evaluation Metrics}
Numerous studies have demonstrated that performing a simple open-loop evaluation is insufficient to fully assess a model's performance~\cite{li2024ego,zhai2023rethinking}, while closed-loop evaluations are hindered by high computational costs and discrepancies between the simulator and the real word. 
NAVSIM provides an intermediary solution between these two evaluation paradigms by introducing the Predictive Driving Model Score (PDMS)~\cite{dauner2024navsim}, which shows a high correlation with closed-loop metrics. 
The PDMS is computed based on five indicators: No-Collision (NC), Drivable Area Compliance (DAC), Time-to-Collision (TTC), Comfort (C), and Ego Vehicle Progress (EP).

In addition to PDMS, NAVSIM also provides an extended benchmark, the Extended Predictive Driving Model Score (EPDMS)~\cite{li2025hydra++}. 
The extended score introduces two new weighted metrics (Lane Keeping, LK, and Extended Comfort, EC), two new multiplicative metrics (Driving Direction Compliance, DDC, and Traffic Light Compliance, TLC), along with a false alarm penalty filtering process.

\vspace{-0.3cm}
\subsection{Implementation Details}

We employ the Transfuser~\cite{chitta2022transfuser} as our perception network, with ResNet34 as the backbone for feature extraction. 
The perception inputs consist of concatenated images from the front-left, front, and front-right cameras, as well as point cloud data covering a \(64\,\text{m} \times 64\,\text{m}\) area. 
In the autoregressive planning module that incorporates MOE, we configure \(E_{private}=5\) domain-specific experts along with \(E_{shared}=1\) shared expert. 
During forward propagation, the top \(K=2\) experts with the highest score are selected for activation. 
The model is trained on the navtrain split using two A100 GPUs with a batch size of 128. 
The initial learning rate is set to \(2 \times 10^{-4}\) and the weight decay is set to \(1 \times 10^{-4}\). 
The model performs 8 autoregressive steps, with each step outputting one trajectory point comprising the \(x\), \(y\) and heading. 
After all autoregressive steps, the generated trajectory is refined to form a 4-second planning trajectory at 2 Hz. 
During the training stage of the perception module, the loss weight coefficients \(\lambda_{\text{sem}}\), \(\lambda_{\text{class}}\), and \(\lambda_{\text{box}}\) are set to 10, 10, and 5, respectively. 
In the end-to-end training stage, the weight coefficients \(\lambda_{\text{sem}}\), \(\lambda_{\text{class}}\), \(\lambda_{\text{box}}\), \(\lambda_{\text{traj}}\), and \(\lambda_{\text{NLL}}\) are set to 1, 1, 0.5, 15, and 0.2, respectively.

\vspace{-0.3cm}
\subsection{Main Results}

\textbf{Quantitative Results.} On the Navtest benchmark, we benchmarked ARTEMIS against several state‑of‑the‑art methods, the outcomes are summarized in Tab.\ref{tab:navsim_v1}. 
With a ResNet‑34 backbone, ARTEMIS attains 87.0 PDMS on the navtest split, delivering competitive performance across most models.
Notably, ARTEMIS achieves a markedly superior EP , NC and C score, evidencing its strong trajectory planning capability and environmental adaptability.
We further evaluated ARTEMIS on the Navtest benchmark with extended metrics (Tab. \ref{tab:navsim_v1_1}). 
It demonstrates that our model significantly outperforms all other baselines using the same ResNet-34 backbone, achieving state-of-the-art (SOTA) results. 
Notably, our method substantially exceeds comparative methodologies on critical metrics such as TTC and EP.
These results underscore ARTEMIS’s robustness and superior performance across diverse evaluation criteria.

\begin{figure*}[ht!]
    \centering
    \includegraphics[width=1\textwidth]{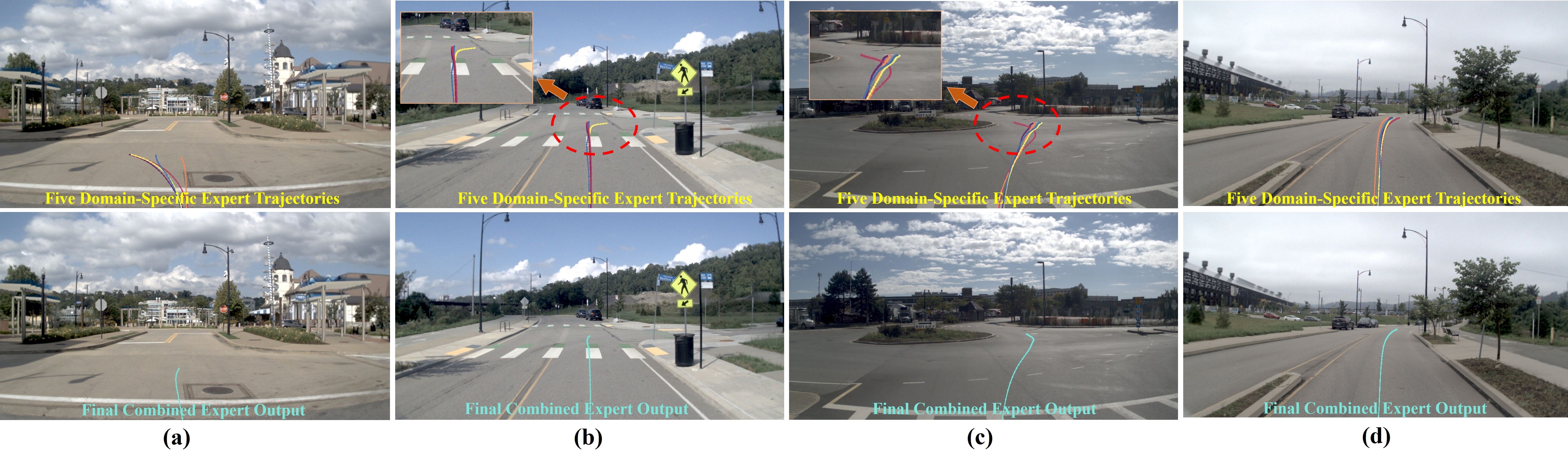}
    \vspace{-0.5cm}
    \caption{Qualitative analysis results of ARTEMIS on the navtest split. The qualitative results demonstrate the differentiated processing strategies of five domain-specific experts and the mixture expert for the same scenario. The visualization trajectories are sampled from 2Hz to 10Hz.}
    \vspace{-0.3cm}
    \label{fig:qualitative}
\end{figure*}

\textbf{Qualitative Results}. Fig.\ref{fig:qualitative} demonstrates four representative driving scenarios sampled from the navtest dataset to qualitatively evaluate the proposed method. To highlight how individual experts respond to the same scenario, trajectories generated by each domain-specific expert are drawn in different colors. 

The first example (Fig.\ref{fig:qualitative}(a)) depicts a scenario where the ego vehicle passes through an intersection, with expert behaviors categorized into either left-turning or proceeding straight. 
The trajectory processed through routing network fusion prioritizes the straight-ahead solution from the orange expert in its planned trajectory.

Fig.\ref{fig:qualitative}(b), as the ego vehicle approaches a T-intersection, the yellow expert opts for a right turn while the remaining experts choose to continue straight ahead. 
The trajectory processed through routing network fusion incorporates elements from the yellow expert's result. 

Fig.\ref{fig:qualitative}(c) and Fig.\ref{fig:qualitative}(d) further illustrate two different scenarios: navigating a roundabout and lane entry selection.
In the Fig.\ref{fig:qualitative}(c) case, the experts' planning proposals show significant divergence: except for the red expert, the others incorrectly handle the scenario by simply choosing to drive toward the right-front direction, whereas the red expert successfully captures the roundabout road features and correctly navigates around it. 
Notably, the final planning trajectory using the routing network for expert fusion primarily references the red expert's solution, demonstrating the effectiveness and rationality of the intrinsic routing mechanism.
The visualization of the lane entry selection scenario (Fig.\ref{fig:qualitative}(d)) also clearly reveals different experts' preferences for different entrance lane positions in the current context. 

\begin{figure*}
    \centering
    \includegraphics[width=1\textwidth]{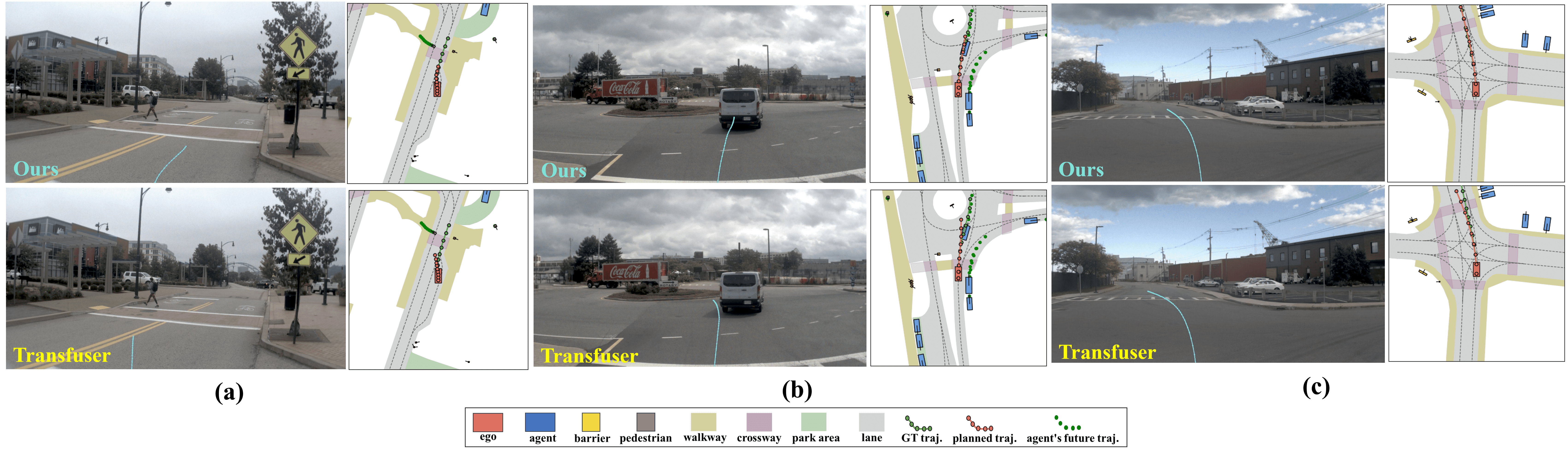}
    \vspace{-0.5cm}
    \caption{Qualitative comparison between ARTEMIS and the baseline Transfuser on the navtest split. The results indicate that ARTEMIS, which employs a mixture of experts model, can more accurately respond to erroneous control commands (Figure (a)), while significantly improving capabilities such as drivable area compliance (Figure (b)) and lane keeping (Figure (c)), demonstrating enhanced trajectory planning stability.}
    \label{fig:qualitative2}
    \vspace{-0.5cm}
\end{figure*}

In addition, we also visually compared the trajectory planning performance of ARTEMIS and Transfuser across different scenarios, with results shown in Fig.\ref{fig:qualitative2}.

\vspace{-0.2cm}
\subsection{Ablation Study}

\textbf{Component‑Wise Impact Analysis.} To assess the contribution of each architectural element, we constructed three ablated variants by individually removing the Autoregressive Planning Module with Mixture-of-Experts (AME), the Mixture‑of‑Experts module (MoE), and the trajectory refinement module (TR). 
The results, summarized in Tab.\ref{tab:ablation_components}, show that eliminating any single component degrades performance, thereby confirming the necessity of all three. 
Specifically, omitting the autoregressive module reduces PDMS by 3.0 points, indicating that the autoregressive paradigm is essential for capturing temporal dependencies among waypoints and for accurately accommodating evolving environmental context. Removing the MoE module yields a PDMS drop of 4.1 points, underscoring the advantage of the MoE architecture in dynamically adapting to diverse driving scenarios and behaviour patterns. 
Finally, excising the cascaded‑refinement module lowers PDMS by 2.3 points, demonstrating that this stage effectively mitigates sampling instabilities that can arise during autoregressive trajectory generation.
\begin{table}[ht]
  \centering
   \vspace{-0.3cm}
  \caption{Ablation Study of Model Components.}
  \label{tab:ablation_components}
  \begin{tabular}{ccc|ccccc>{\columncolor{gray!20}}c}
    \toprule
    \textbf{AME} & \textbf{MOE} & \textbf{TR} &
    \textbf{NC} & \textbf{DAC} &
    \textbf{EP} & \textbf{TTC} &
    \textbf{C} & \textbf{PDMS} \\
    \midrule
    $\times$ & $\checkmark$ & $\checkmark$ & 97.8 & 92.3 & 78.9 & 93.4 & 99.9 & 84.0 \\
    $\checkmark$ & $\times$ & $\checkmark$ & 97.6 & 91.7 & 78.2 & 92.7 & 99.8 & 82.9 \\
    $\checkmark$ & $\checkmark$ & $\times$ & 97.7 & 93.1 & 79.7 & 93.3 & 99.9 & 84.7 \\
    $\checkmark$ & $\checkmark$ & $\checkmark$ &
    \textbf{98.3} & \textbf{95.1} & \textbf{81.4} & \textbf{94.3} & \textbf{100} & \textbf{87.0} \\
    \bottomrule
  \end{tabular}
\vspace{-0.3cm}
\end{table}

\textbf{The Routing Network.} To verify the effectiveness of the intrinsic routing mechanism, we compare it with an explicit driving‑command guidance scheme. 
The experimental results in Tab.\ref{tab:expert_assignment} show that the intrinsic routing mechanism effectively prevents inappropriate assignment of domain‑specific experts when explicit guidance deviates from the actual scene.

We also conducted experiments by fixing the activation to single experts within the model.
The results presented in Table.\ref{tab:single_expert_activation} demonstrate that activating any individual expert in isolation fails to achieve optimal performance, underscoring the necessity of the routing network for facilitating effective expert allocation. 
Moreover, the results confirms that the model training process did not exhibit disproportionate preference toward utilizing any particular expert, indicating balanced expert utilization throughout the learning process.
\begin{table}[ht]
  \centering
  \vspace{-0.3cm}
  \caption{Comparison of Expert‑Assignment Strategies.}
  \label{tab:expert_assignment}
  \begin{tabular}{l|ccccc>{\columncolor{gray!20}}c}
    \toprule
    \makecell{Expert \\ Assignment} & \textbf{NC} & \textbf{DAC} &
    \textbf{EP} & \textbf{TTC} & \textbf{C} &
    \textbf{PDMS} \\
    \midrule
    \makecell{Driving Cmd.}            & 97.8   & 92.1   & 78.5   & 93.0   & 99.8  & 83.5   \\
    \makecell{Intrinsic Routing} & \textbf{98.3} & \textbf{95.1} & \textbf{81.4} & \textbf{94.3} & \textbf{100} & \textbf{87.0} \\
    \bottomrule
  \end{tabular}
\vspace{-0.1cm}
\end{table}

\begin{table}[ht]
\centering
\vspace{-0.5cm}
\caption{Performance of Single Expert Activation Strategies}
\label{tab:single_expert_activation}
\begin{tabular}{c|ccccc>{\columncolor{gray!20}}c}
\toprule
\makecell{Individually \\Activated Expert} & \textbf{NC} & \textbf{DAC} &
    \textbf{EP} & \textbf{TTC} & \textbf{C}&\textbf{PDMS} \\
\midrule
Expert 1 & 97.5 & 92.5 & 79.0 & 92.2 & 99.9 & 83.4 \\
Expert 2 & 98.0 & 94.7 & 81.0 & 93.7 & 100 & 86.2 \\
Expert 3 & 97.4 & 91.0 & 77.4 & 91.9 & 99.8 & 81.9 \\
Expert 4 & 97.8 & 93.8 & 80.0 & 93.3 & 99.9 & 85.1 \\
Expert 5 & 98.1 & 94.1 & 80.2 & 93.2 & 100 & 85.4 \\
\midrule
Intrinsic Routing & \textbf{98.3} & \textbf{95.1} & \textbf{81.4} & \textbf{94.3} & \textbf{100} & \textbf{87.0} \\
\bottomrule
\end{tabular}
\vspace{-0.1cm}
\label{tab:expert_performance}
\end{table}

\textbf{Cascaded Refinement Layers.} Tab.\ref{tab:refinement_layers} reports how varying the number of refinement layers influences model performance. 
Increasing the cascade depth improves performance up to two layers, beyond this point, the gain stabilises and the marginal benefit diminishes.
\begin{table}[ht]
  \centering
    \vspace{-0.3cm}
  \caption{Impact of the Number of Refinement Layers.}
  \label{tab:refinement_layers}
  \begin{tabular}{c|ccccc>{\columncolor{gray!20}}c}
    \toprule
    \makecell{Refinement\\Layers} & \textbf{NC} & \textbf{DAC} &
    \textbf{EP} & \textbf{TTC} & \textbf{C} &
    \textbf{PDMS} \\
    \midrule
    1 & 98.1   & 93.9   & 80.0   & \textbf{94.3}   & \textbf{100}  & 85.7   \\
    2 & \textbf{98.3} & \textbf{95.1} & \textbf{81.4} & \textbf{94.3} & \textbf{100} & \textbf{87.0} \\
    3 & 97.9   & 93.7   &80.6   & 93.2   & 99.9  & 85.4   \\
    \bottomrule
  \end{tabular}

\end{table}

\textbf{Domain‑Specific Experts.} Tab.\ref{tab:num_experts} investigates the impact of varying the number of experts. 
Enlarging the expert pool from three to five gradually increases the model’s performance, indicating enhanced capability for handling complex scenarios. 
However, expanding to ten experts leads to a performance drop of 1.5 points, suggesting that, given limited training data, an excessive number of experts disperses resources and causes functional overlap.
\begin{table}[ht]
  \centering
  \vspace{-0.1cm}
  \caption{Effect of the Number of Domain‑Specific Experts.}
  \label{tab:num_experts}
  \begin{tabular}{c|ccccc>{\columncolor{gray!20}}c}
    \toprule
    \makecell{Domain‑Specific\\ Experts} & \textbf{NC} & \textbf{DAC} &
    \textbf{EP} & \textbf{TTC} & \textbf{C} &
    \textbf{PDMS} \\
    \midrule
    3  & 98.2   & 94.1   & 80.5   & 94.1   & 99.9  & 85.9   \\
    5  & \textbf{98.3} & \textbf{95.1} & \textbf{81.4} & \textbf{94.3} & \textbf{100} & \textbf{87.0} \\
    10 & 98.1   & 93.6   & 80.0   & 93.9   & \textbf{100}  & 85.5   \\
    \bottomrule
  \end{tabular}
\vspace{-0.3cm}
\end{table}

\textbf{Batch Reallocation on Training Speed.} To assess the contribution of batch reallocation, we compared training speed with and without this strategy across different batchsize.
As presented in Tab.\ref{tab:batch_reordering}, under identical hardware conditions, batch reallocation significantly accelerates training, the training samples per second from 19.2 to 43.5 as the batch size grows from 64 to  256. Although reallocation introduces extra overhead, it is negligible relative to expert network computation and is offset by the gain in parallel efficiency.
\begin{table}[ht]
  \centering
  \vspace{-0.5cm}
  \caption{Effect of Batch Reallocation on Training Samples per Second under Different Batchsizes (Using a Single NVIDIA A100 GPU).}
  \label{tab:batch_reordering}
  \begin{tabular}{c|ccc}
    \toprule
    \textbf{Batch Reallocation} & \textbf{64} & \textbf{128} & \textbf{256} \\
    \midrule
    $\checkmark$ & 19.20 & 44.81 & 43.54 \\
    $\times$     & 2.31 & 1.95 & 1.61 \\
    \midrule
    \textbf{Training Speedup Factor$\uparrow$} & 7.31 & 21.97 & 26.2 \\
    \bottomrule
  \end{tabular}
\vspace{-0.5cm}
\end{table}

\section{Conclusion}
This paper introduces ARTEMIS, Autoregressive End-to-End Trajectory Planning with Mixture of Experts for Autonomous Driving. 
In contrast to conventional static paradigms that synthesize complete trajectories in one-shot inference paradigms, ARTEMIS implements a sequential decision-making process that enables modeling of trajectory evolution. 
Through its integrated Mixture-of-Experts architecture with dedicated routing networks, ARTEMIS dynamically captures the intrinsic dynamic characteristics of driving behavior and effectively accommodates diverse driving environments. 
Extensive quantitative evaluations conducted on the NAVSIM benchmark demonstrate that ARTEMIS achieves highly competitive performance. 
Given its flexibility and adaptability, this framework exhibits substantial deployment potential in complex scenarios and establishes a promising direction for future autonomous driving research.

\bibliography{ral2025_xn}

\end{document}